%% file: main.tex
\documentclass{article}
\usepackage{arxiv}
\usepackage[utf8]{inputenc}
\usepackage[T1]{fontenc}
\usepackage{natbib}
\usepackage[colorlinks=true,linkcolor=blue!50!black,citecolor=blue!50!black,urlcolor=blue!50!black,pdftitle={DROD: Dual-Ratio Ordered Dropout for Federated Continual Learning on Heterogeneous Devices},pdfauthor={Berk Sezer, Ozgur Soysal}]{hyperref}
\usepackage{url}
\usepackage{booktabs}
\usepackage{amsfonts}
\usepackage{amsmath}
\usepackage{amsthm}
\usepackage{amssymb}
\usepackage{graphicx}
\usepackage{xcolor}
\usepackage{subcaption}
\usepackage{float}
\usepackage{placeins}

\date{}
\title{DROD: Dual-Ratio Ordered Dropout for Federated Continual Learning on
Heterogeneous Devices}

\renewcommand{\shorttitle}{DROD: Dual-Ratio Ordered Dropout}
\renewcommand{\headeright}{}
\renewcommand{\undertitle}{}
\author{
  Berk~Sezer \\
  Middle East Technical University \\
  \texttt{berk.sezer@metu.edu.tr}
  \And
  Özgür~Soysal \\
  Stanford University \\
  \texttt{soysal@stanford.edu}
}
\begin{document}

\maketitle

\begin{abstract}
Federated learning for capacity-heterogeneous clients and federated
continual learning are each well studied on their own, but their
intersection -- clients holding differently sized models while every
client's data stream also drifts over time -- remains underexplored. We study this
setting directly, training a ResNet-18 fleet from scratch on Split-CIFAR-100
(10 tasks, class-incremental), and introduce DROD (Dual-Ratio Ordered
Dropout), which builds on FjORD's Ordered Dropout by using two independent
reduction axes to extract sub-models: a shortcut ratio and a branch ratio.
On a shared parameter/FLOP budget ladder, DROD achieves the highest mean
accuracy among the heterogeneous-model federated learning strategies we
compare. Its advantage over the two strongest baseline geometries persists
across every replay-buffer size we test and under ER, DER++, and BiC; under
EWC, all three geometries collapse alike.
\end{abstract}

\section{Introduction}
\label{sec:intro}

Model-heterogeneous federated learning (FL) tackles device disparity by allocating capacity-tailored submodels to participating clients, using channel-width slicing~\citep{diao2021heterofl,alam2022fedrolex,horvath2021fjord}, depth scaling~\citep{kim2023depthfl}, joint width-depth allocation~\citep{ilhan2023scalefl}, or importance-aware pruning~\citep{wu2024fiarse}. However, these approaches are evaluated almost exclusively on stationary, single-task distributions. Conversely, federated continual learning (FCL) studies non-stationary task streams, but predominantly under the assumption of identical client architectures. The intersection of these two settings -- continual learning across compute-heterogeneous edge fleets -- remains underexplored. Existing methods in this setting~\citep{feddcl2025,fedlscl2025} avoid direct weight aggregation, relying instead on knowledge distillation into a server model with a frozen, pre-trained backbone. FedDCL~\citep{feddcl2025} does train from scratch, but only in its grayscale benchmark, where both the server and the client models are two- to four-layer CNNs over MNIST, EMNIST, and Fashion-MNIST -- a considerably lighter regime than a full ResNet-18 on CIFAR-100; its own CIFAR-100 experiments freeze the server Vision Transformer and train only the classifier head. In contrast, we directly aggregate weights across heterogeneous client capacities and train the full ResNet-18 architecture entirely from scratch on Split-CIFAR-100.

These model-heterogeneous FL methods differ mainly in which structural dimension a smaller client cuts and how its submodel is chosen: a fixed contiguous band of channels~\citep{diao2021heterofl}, a band rotated across channel indices between rounds so that every parameter is eventually trained by every tier~\citep{alam2022fedrolex}, a joint width--depth split~\citep{ilhan2023scalefl}, whole layers rather than channels~\citep{kim2023depthfl}, or a magnitude-selected weight mask~\citep{wu2024fiarse}. FjORD~\citep{horvath2021fjord} differs in kind rather than degree: instead of fixing one submodel per client, it resamples a width rate at every local step (capped by the client's capacity), so training itself sweeps a family of nested widths and induces an importance ordering on the channels -- \emph{Ordered Dropout}. What all of these share is a single reduction axis and a scalar notion of ``capacity''. We observe that a residual block instead exposes two independent reduction ratios: a \emph{shortcut ratio} $s$ on the block's carried stream and a \emph{branch ratio} $b$ on its residual transform. Hence, we propose Dual-Ratio Ordered Dropout (DROD), which uses both ratios at once rather than a single width.

\section{Method: Dual-Ratio Ordered Dropout (DROD)}
\label{sec:method}

\paragraph{Background: Ordered Dropout (OD).}
  FjORD samples a width $p$, capped by each client's capacity, at every local
  iteration.\footnote{With tiers $\{0.25,0.5,1.0\}$, a capacity-$1.0$ client
  samples from all three widths, whereas a capacity-$0.5$ client samples only
  from $\{0.25,0.5\}$.}
  Retaining contiguous channel prefixes makes narrower submodels nested within
  wider ones and trains low-index channels more frequently, inducing an
  importance ordering.

\begin{figure}[t]
\centering
\begin{subfigure}[b]{0.32\linewidth}
\centering
\includegraphics[trim=40 135 43 190,clip,width=\linewidth]{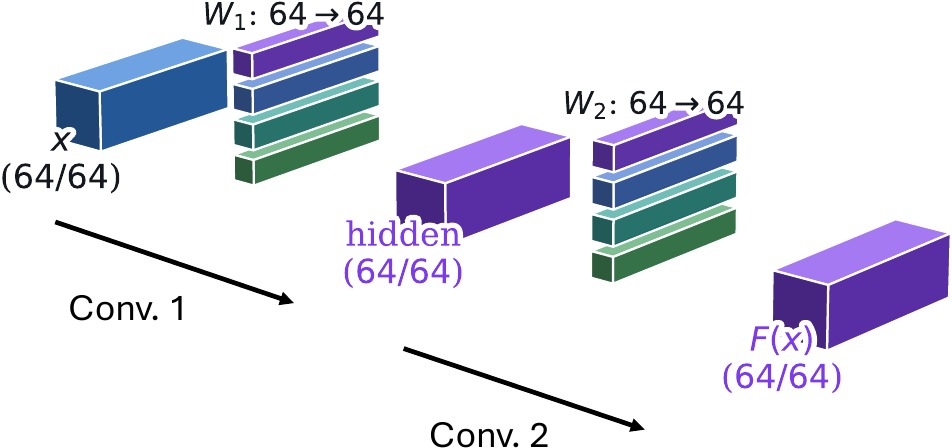}
\caption{Full block ($s{=}1$, $b{=}1$)}
\end{subfigure}
\hfill
\begin{subfigure}[b]{0.32\linewidth}
\centering
\includegraphics[trim=40 135 43 190,clip,width=\linewidth]{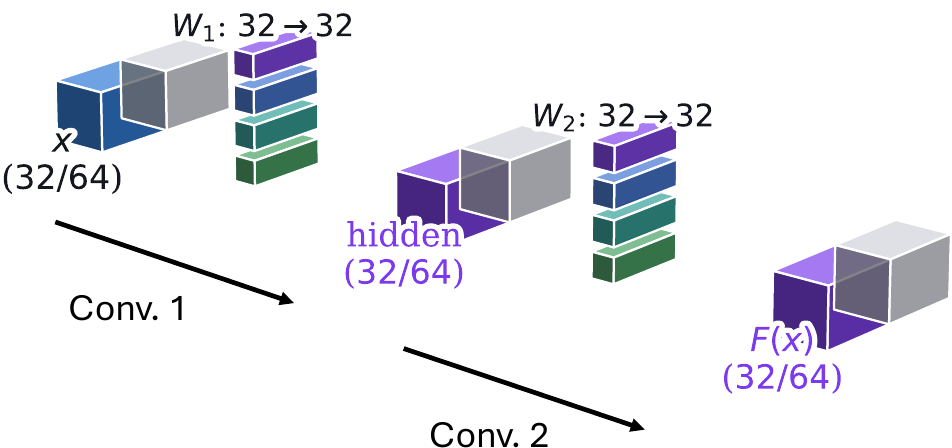}
\caption{Width slicing / FjORD ($\theta{=}0$)}
\end{subfigure}
\hfill
\begin{subfigure}[b]{0.32\linewidth}
\centering
\includegraphics[trim=40 135 43 190,clip,width=\linewidth]{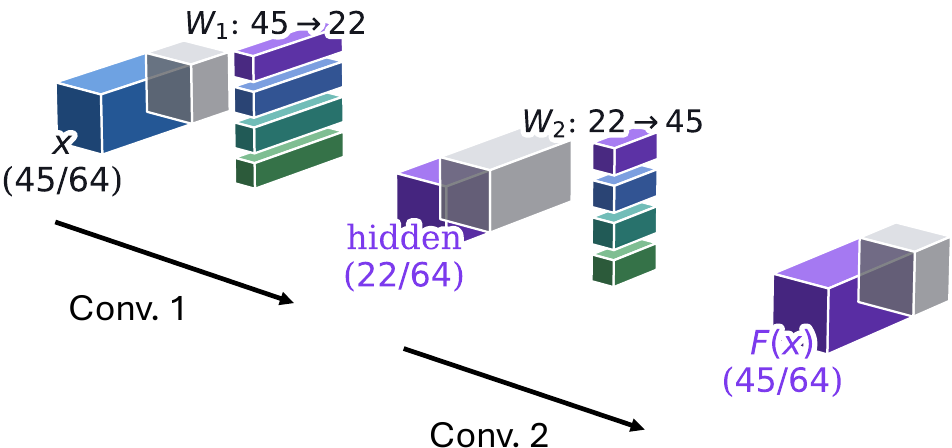}
\caption{DROD's interior ($\theta{=}0.5$)}
\end{subfigure}
\caption{One residual block under three configurations, $C{=}64$; in every
panel each tensor is drawn at its full capacity, with the channels that tier
keeps in color and the dropped remainder in gray, and each label reading
kept/total. (a) The
full, unreduced block, shown as a baseline; (b) width slicing / FjORD,
where the shortcut and the branch narrow together to the same width; (c)
DROD's interior, where the shortcut stays near-full-width ($s{=}0.70$) and
only the branch narrows to a low rank ($b{=}0.49$) -- a low-rank transform
on a near-full-width shortcut, decoupled rather than welded. Shortcut-tiering,
the precursor technique that pushes $b\to0$, needs no diagram of its own: it
collapses the branch to $F(x){=}0$, leaving the shortcut passed through
unchanged.}
\label{fig:block-3d}
\end{figure}

\paragraph{Two independent ratios.} DROD uses two ratios: a \textbf{shortcut ratio} $s \in (0,1]$ sets the width of $x$, \textbf{branch ratio} $b \in (0,1]$ sets the width of $F(x)$'s hidden layer\footnote{A residual block computes
$y = \mathrm{ReLU}(x + F(x))$ on a $C$-channel stream $x$, where $F$ is the
block's own two-convolution transform.} as a fraction of the
already-$s$-narrowed stream, so the block's two convolutions
become $W_1 : sC \to bsC$ and $W_2 : bsC \to sC$.
Figure~\ref{fig:block-3d} shows the two ratios acting on one block, and how
DROD sets them apart from width slicing. Setting $b{=}1$ recovers
FjORD's block exactly: the hidden layer is then as wide as the stream, and
$s$ alone controls the block. At $s{=}1$ the block reduces to nesting the
hidden width alone, a construction that
MatFormer~\citep{devvrit2024matformer} applies to the transformer
feed-forward block.

Every weight in the block is sized by $(s,b)$ alone. A $k \times k$
convolution with $C_{\mathrm{in}}$ input and $C_{\mathrm{out}}$ output
channels has $k^2 C_{\mathrm{in}} C_{\mathrm{out}}$ parameters, so summing
$W_1$ ($sC \to bsC$) and $W_2$ ($bsC \to sC$) gives
\begin{equation}
\mathrm{params}(\mathrm{block}) = 2k^2\,b\,s^2\,C^2.
\label{eq:block-params}
\end{equation}
At fixed spatial resolution, FLOPs scale with the same $bs^2$ factor.
The shortcut ratio $s$ affects the channel dimensions of both sides of
each convolution and therefore enters quadratically, whereas the branch
ratio $b$ controls only the intermediate width and therefore enters
linearly.
FjORD reports the same behavior for its single rate -- FLOPs and
parameters there fall as $1/p^2$ at width $p$ -- DROD simply separates the
two powers of that exponent into two independently settable ratios instead
of coupling them into one.

\paragraph{The precursor: shortcut-tiering.} As $b \to 0$, the residual branch $F(x)$ vanishes, leaving the block as an exact identity.\footnote{Since $x$ is the output of the preceding block's ReLU, it is already non-negative, so $y=\mathrm{ReLU}(x)=x$ exactly.}
Before deriving DROD, we tested this limiting case on its own as shortcut-tiering where blocks are reduced to exact identities (deepest-first within each stage) while every tier retains the full shortcut pathway and shares a single classifier head. Shortcut-tiering outperformed plain (FedAvg) width slicing at both buffer totals where we ran that comparison ($200$ and $12{,}800$). Its limitation, however, is that it can only remove entire residual branches, placing a hard ceiling on the amount of model reduction it can provide.\footnote{Only one block per stage is
droppable, and the last stage alone holds 75\% of the network's
parameters, so dropping every droppable block still leaves 44\% of
parameters standing, meaning shortcut-tiering cannot reach, on its
own, the $16\times$ fleet spread required by a 0.25-tier client, needing a $1/0.25^2=16$ times reduction}.

\paragraph{The interior: DROD.} Width slicing ($b{=}1$) and the
shortcut-tiering limit ($b{\to}0$) bound the broader two-ratio design space.
DROD explores it by varying $b \in (0,1]$ together with $s \in
(0,1]$. By Equation~\eqref{eq:block-params},
$\mathrm{params}(\mathrm{block})$ is proportional to $bs^2$, so fixing the
block's budget at a fraction $\rho$ of the full block's parameters
constrains $(s,b)$ to the curve $bs^2 = \rho$, not a single point
(see Appendix Figure~\ref{fig:theta-curve}). We parameterize that curve by $\theta \in
[0,1]$:
\[
s = \rho^{(1-\theta)/2}, \qquad b = \rho^{\theta}.
\]
$\theta{=}0$ is width slicing; $\theta{=}1$ is the other endpoint of this
particular nonzero-budget curve, a full-width shortcut with branch ratio
$b{=}\rho$. It does not reach shortcut-tiering, which lies at $b{=}0$ and is
recovered only as a limiting case when the branch budget vanishes. At our
mid tier ($\rho{=}0.2514$), $\theta{=}0.5$ lands at
$s{=}0.70$, $b{=}0.49$ (Figure~\ref{fig:block-3d}(c)).\footnote{We solve
$(s,b)$ against the whole network rather than the block identity above: $s$
snaps to a multiple of $1/64$ so that every stage width stays integral, and
$b$ is then fit to the measured parameter ratio, since the stem, the
downsample convolutions and the classifier ($2.0\%$ of the model) lie
outside the branch axis. Hence $0.70/0.49$ rather than the closed form's
$0.71/0.50$; the quoted pair is the one the runs use.}

\paragraph{Why $\theta$ is a good axis to sweep.}
  Varying $\theta$ changes how a fixed budget is divided between the shortcut
  and residual branch. Across the sweep, parameters and inference FLOPs remain
  matched within $0.5\%$, so the same configurations satisfy both budgets.
  Moreover, GroupNorm keeps the branch-output magnitude nearly constant as $b$
  shrinks: a $16\times$ reduction in branch width changes $\lVert F(x)\rVert$
  by only $0.05\%$. Thus, $\theta$ primarily controls the branch's rank rather
  than its magnitude.

\textbf{DROD} (Dual-Ratio Ordered Dropout) is Ordered Dropout on the shortcut (s) and branch (b) ratios at once. At every local iteration, a client samples a nested $(s,b)$
pair from the subset of fractions its capacity can afford (i.e. points satisfying
$s \le s_{\mathrm{own}}$ and $sb \le s_{\mathrm{own}} b_{\mathrm{own}}$,
where $(s_{\mathrm{own}}, b_{\mathrm{own}})$ is the client's own tier
ceiling) exactly mirroring FjORD's per-iteration resampling of $p$, only
over two coupled coordinates instead of one. This preserves the nesting
property OD's importance ordering depends on: a lower-tier client's sampled
block is always a prefix of a higher-tier client's along both the
shortcut and the branch, so the server retains the same prefix-sliced
averaging rule used by FjORD.

\paragraph{Relation to prior work.}
  Like FjORD and other nested-network methods, DROD trains submodels of different
  sizes within a shared model. Its closest structural analogue is MatFormer,
  which varies a Transformer's feed-forward hidden width while keeping its input
  and output dimensions fixed. DROD brings the same separation to residual
  networks, allowing clients to reduce the residual branch without narrowing the
  shortcut by the same amount.

\section{Experimental setup}
\label{sec:setup}

\paragraph{Task stream.} We use Split-CIFAR-100: the 100 classes are cut, in
a fixed order shared across all seeds, into $T{=}10$ disjoint tasks of 10
classes, reaching every client simultaneously at synchronous boundaries. Training is fully class-incremental, using a single shared 100-way head with unmasked cross-entropy and no task identifier during either training or testing. Each task
contributes roughly 5{,}000 training images and is trained for 25
communication rounds of 4 local epochs, so a run spans 250 rounds.

\paragraph{Federation and fleet.} Ten clients participate in every round.
Within each task's ten classes, we split samples across clients with a
Dirichlet($\alpha{=}0.5$) label skew, redrawn fresh for every task rather
than fixed once for the whole run, so a client over-represented in one class
this task need not stay that way next task. Each
client holds on the order of 500 images of the current task. Capacity tiers are assigned round-robin across clients (4/3/3 per tier) once at the start of each run and remain fixed throughout, reflecting persistent device-level hardware heterogeneity. The server aggregates client updates coordinate-by-coordinate,
weighted by sample count; a coordinate that only the wider tiers
instantiate is averaged over just those clients that hold it.

\paragraph{Model and shared budget.} The backbone is a ResNet-18 with a CIFAR
stem, trained from scratch, using GroupNorm in place of BatchNorm
throughout. We adopt GroupNorm~\citep{wu2018groupnorm} following
\citet{hsieh2020quagmire}, who isolate normalization as a primary cause of the
accuracy that decentralized learning loses under skewed label partitions\footnote{BatchNorm's minibatch statistics
diverge across clients holding different label distributions, so averaging
them is not a meaningful operation, whereas GroupNorm is independent of a
client's label distribution and the trade favors us under skew.
Appendix~\ref{app:groupnorm} gives the numbers and the caveats.}. At full
capacity the model holds $11.22$M parameters and costs $1.11$ GFLOPs per
inference. Every method (variant) we compare places its three tiers on one shared
budget ladder: $1.000 / 0.251 / 0.064$ ratios of the full ResNet18 model's parameters or FLOPs. \footnote{Each
tier's parameter and FLOP counts are measured directly, by rebuilding that
tier and profiling it rather than trusting the configuration's arithmetic.
On whichever metric a grid is matched on, every row lands within $2.3\%$
(parameters) or $4.4\%$ (FLOPs) of its rung, and the arms that decide the
ranking are tighter still: Width~$\times$~FjORD sits on both ladders
exactly, and DROD $\theta{=}0.50$ within $0.5\%$. The \emph{unmatched}
metric is deliberately left free, and that is where the families separate.
Holding FLOPs fixed drives a depth-tiered fleet down to $3\%$ of the
reference model's parameters at its mid tier, while holding parameters
fixed lets the same fleet spend $37\%$ of its FLOPs there; FIARSE's
magnitude mask deletes weights without removing a single multiply, so it
buys communication and no compute. ScaleFL is the one arm on neither
ladder, since it fixes its own width/depth split: its mid tier spends
$65\%$ of the reference model's parameters where every other row spends
about $25\%$.}

\paragraph{Continual-learning mechanism.} Every experiment in this paper is federated continual learning, so it carries both a federated-learning axis and a continual-learning axis. Since our method's novelty is on the model-heterogeneous federated-learning side rather than a new continual-learning objective, we prioritized breadth on FL strategies over breadth on CL algorithms. Experience Replay (ER) is therefore our default anti-drift mechanism throughout: it is both the canonical baseline and the strongest performer we measured in this setting. To check that our findings hold beyond ER, Table~\ref{tab:rq3-generality} evaluates the top three method families from Table~\ref{tab:rq3-main} (DROD, Width~$\times$~FjORD, and Shortcut~$\times$~FjORD) under four different continual-learning objectives; the per-tier breakdown is in Table~\ref{tab:rq3-clmethod}. Under ER, each client keeps a local reservoir buffer whose samples are appended to every local mini-batch. We evaluate four fleet-wide buffer totals, 200, 3{,}200, 6{,}400 and 12{,}800 exemplars, distributed uniformly across the fleet (each client gets the total exemplars divided by client count) unless stated otherwise. The main grid uses 6{,}400; Appendix~\ref{sec:hparams} explains the choice of totals.

\paragraph{Metrics.} Both readouts come off the identical forward passes and
therefore describe the same weights: \textbf{class-IL} accuracy takes the
argmax over all 100 classes (chance $0.010$) and \textbf{Task-IL} accuracy
restricts the argmax to the scored task's own 10 (chance $0.100$), the
standard \texttt{mask\_classes} protocol. Class-IL is the primary metric since in real deployment there is no task oracle. Defining $R_{t,i}$ as accuracy on
task $i$ after training through task $t$, we report ACC as the mean of the
final row, $\frac{1}{T}\sum_i R_{T-1,i}$, and average forgetting
\citep{chaudhry2018forgetting} as the mean over past tasks of
$\max_{s<T-1} R_{s,i} - R_{T-1,i}$. A tier's score is the shared model
evaluated at that tier's own extracted width; \emph{fleet} denotes the
unweighted mean over the three tiers, formed per seed and then averaged.

\paragraph{Seeds, hardware, and error bars.} All reported intervals are
$\pm$ one sample standard deviation over $n=3$ seeds. Appendix~\ref{sec:hparams} gives the
complete hyperparameter list and states the three budget axes that this
protocol deliberately does \emph{not} hold matched.

\section{Results}
\label{sec:results}

\input{rq3-tables}

Every row in Table~\ref{tab:rq3-main} pairs a \emph{reduction axis}, which
structural dimension of the ResNet a smaller tier actually cuts (its width,
its depth, or the shortcut/branch ratio DROD interpolates between), with an
\emph{aggregation rule}, the policy that decides which slice of the model a
client trains each round and how the server folds tiers of different sizes
back into one shared model (a fixed slice per tier for the FedAvg-style
baselines, FjORD's per-iteration resampled slice, or one of the other
schemes from prior work).\footnote{The table is not a full $m \times n$
grid: FedRolex~\citep{alam2022fedrolex} and
ScaleFL~\citep{ilhan2023scalefl} are each defined for a single reduction
axis, so each appears once rather than once per axis.
Appendix~\ref{app:grid-shape} gives the explanation in full.} Every
row sits at the same shared inference-FLOP budget, so the ranking is a
statement about mechanism, not about spend. We match on FLOPs here because
that is the budget a client actually spends to run its model; the
parameter-matched grid --- the convention the model-heterogeneous
federated-learning literature inherited from its communication-cost
motivation --- is Table~\ref{tab:rq3-grid-params-6400}, and the two agree on the
six leading rows, differing mainly in where the depth family lands. For the
arms that decide the ranking, then, the choice of budget is immaterial, and
for a structural reason rather than a numerical coincidence: narrowing a
ResNet's channels cuts its parameters and its multiplies in step, so one
configuration meets both targets at once and every width-family and DROD row
is the same run in both tables. DROD $\theta{=}0.50$ reads $0.3125$ and
Width~$\times$~FjORD $0.2741$ under either budget. Only the depth and
shortcut axes have to be re-solved per ladder, and two groups move
materially: the depth rows fall by $0.028$ to $0.044$ once made to pay for
their compute rather than only for their weights, and
Shortcut~$\times$~FIARSE by $0.051$, because a magnitude mask can spend part
of a parameter budget but none of a FLOP budget, leaving the structural cut
to absorb all of it. Every other row moves by $0.008$ or less. This
table fixes the continual-learning method at plain experience replay and
varies the reduction axis and the aggregation rule; Table~\ref{tab:rq3-generality}
does the opposite, fixing the three leading reduction-axis/aggregation-rule
combinations from Table~\ref{tab:rq3-main} and varying the continual-learning method instead.

\paragraph{DROD has the highest mean in the matched-budget grid.} DROD at
$\theta{=}0.50$ has the highest mean on both readouts among all seventeen arms.
Against every one of the other sixteen, the lead also holds in every one of the
three seeds, not just on average. The strongest arm outside our own family is
Width~$\times$~FjORD at $0.2741$ class-IL: the gap of $0.0383$ is larger than
each arm's own sample standard deviation ($0.0160$ for DROD,
$0.0084$ for Width~$\times$~FjORD), so their $\pm1$ sd intervals do not overlap.

\input{rq3-generality}

\paragraph{The ordering holds under other continual-learning methods.}
Table~\ref{tab:rq3-generality}
highlights that DROD retains the highest mean under DER++ and BiC, exceeding the next-best geometry by 0.021 and 0.026, respectively, with the lead holding in every seed. All geometries collapse under EWC, where DROD is still highest but the three are separated by only 0.004.\footnote{For each baseline family, we retain its stronger arm from the matched-budget grid. The non-distilled variants outperform their self-distilled counterparts: $0.2741$ versus $0.2719$ for Width~$\times$~FjORD and $0.2518$ versus $0.2368$ for Shortcut~$\times$~FjORD.}

\paragraph{Gains across individual client tiers as well as fleet mean.} A fleet-mean gain can in principle be bought at one tier's expense. It is not here: DROD's high, mid and low tier each score above the corresponding tier of \emph{every} baseline geometry, in every cell we report --- both budget ladders, both replay-buffer totals, the buffer sweep, and all four continual-learning objectives --- without exception, and the margin is generally widest at the lowest-capacity tier. Per-tier tables are in Appendix~\ref{app:full-results} (Tables~\ref{tab:rq3-grid-params-6400}--\ref{tab:rq3-grid-flops-200}, \ref{tab:rq3-clmethod} and~\ref{tab:rq3-bufp-full}).

\begin{figure}[H]
\centering
\includegraphics[width=0.66\linewidth]{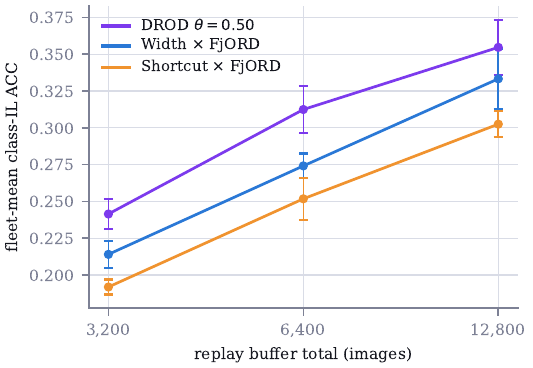}
\caption{Fleet-mean class-IL accuracy against replay-buffer total, uniform
per-tier allocation (every tier gets the same buffer); $n{=}3$, error bars
$\pm1$ sample sd. DROD is above both baselines at every buffer total.
Per-cell values: Table~\ref{tab:rq3-bufp}.}
\label{fig:rq3-bufp}
\end{figure}

\paragraph{Buffer capacity.} The buffer is the other resource a
heterogeneous fleet has to divide. Figure~\ref{fig:rq3-bufp} sweeps the
buffer total over $\{3{,}200,\,6{,}400,\,12{,}800\}$ under a uniform
per-tier allocation. DROD leads at all three totals, by $+0.021$ to
$+0.038$ over the best competing geometry, and the margin holds even at the
smallest buffer, where every geometry has the least to work with.

\input{conclusion}

\bibliographystyle{plainnat}
\bibliography{refs}

\input{appendix}

\end{document}

%% file: rq3-tables.tex
\begin{table}[t]
\centering
\scriptsize
\caption{Reduction axis and aggregation rule at a matched inference-FLOP budget on Split-CIFAR-100 with a fleet-wide replay buffer of 6{,}400. Results are mean $\pm$ sample sd over three seeds; accuracies are fleet means across capacity tiers. The last two columns report mid/low-tier parameter and FLOP savings relative to the full model. $^{\dagger}$The parameter and FLOP budgets coincide. $^{\ddagger}$ScaleFL uses its native width/depth split and is not budget matched.}

\label{tab:rq3-main}
\setlength{\tabcolsep}{3pt}
\begin{tabular}{@{}lccccrr@{}}
\toprule
 & \multicolumn{2}{c}{class-IL} & \multicolumn{2}{c}{Task-IL} & \multicolumn{2}{c}{\% saved, mid / low} \\
\cmidrule(lr){2-3}\cmidrule(lr){4-5}\cmidrule(l){6-7}
reduction $\times$ aggregation & ACC & forg. & ACC & forg. & params & FLOPs \\
\midrule
\textbf{DROD} $\theta{=}0.50$\textsuperscript{$\dagger$} & $0.3125 \pm 0.0160$ & $0.251 \pm 0.019$ & $0.6973 \pm 0.0204$ & $0.003 \pm 0.004$ & 74.9 / 93.6 & 74.8 / 93.7 \\
\textbf{DROD} $\theta{=}1.00$\textsuperscript{$\dagger$} & $0.2812 \pm 0.0048$ & $0.276 \pm 0.008$ & $0.6663 \pm 0.0011$ & $0.012 \pm 0.004$ & 75.0 / 93.5 & 75.4 / 94.0 \\
\textbf{DROD} $\theta{=}0.25$\textsuperscript{$\dagger$} & $0.2760 \pm 0.0091$ & $0.277 \pm 0.010$ & $0.6587 \pm 0.0103$ & $0.014 \pm 0.006$ & 74.8 / 93.7 & 74.8 / 93.7 \\
Width $\times$ FjORD\textsuperscript{$\dagger$} & $0.2741 \pm 0.0084$ & $0.263 \pm 0.004$ & $0.6617 \pm 0.0111$ & $0.006 \pm 0.004$ & 74.9 / 93.6 & 74.9 / 93.7 \\
\textbf{DROD} $\theta{=}0.75$\textsuperscript{$\dagger$} & $0.2723 \pm 0.0083$ & $0.283 \pm 0.013$ & $0.6594 \pm 0.0077$ & $0.012 \pm 0.008$ & 74.7 / 93.6 & 74.7 / 93.8 \\
Width $\times$ FjORD w/ KD\textsuperscript{$\dagger$} & $0.2719 \pm 0.0157$ & $0.279 \pm 0.006$ & $0.6550 \pm 0.0261$ & $0.010 \pm 0.005$ & 74.9 / 93.6 & 74.9 / 93.7 \\
Shortcut $\times$ FjORD & $0.2518 \pm 0.0143$ & $0.291 \pm 0.012$ & $0.6391 \pm 0.0096$ & $0.016 \pm 0.001$ & 76.1 / 93.7 & 75.3 / 93.5 \\
Shortcut $\times$ FjORD w/ KD & $0.2368 \pm 0.0097$ & $0.312 \pm 0.008$ & $0.6232 \pm 0.0075$ & $0.023 \pm 0.006$ & 76.1 / 93.7 & 75.3 / 93.5 \\
ScaleFL\textsuperscript{$\ddagger$} & $0.2149 \pm 0.0054$ & $0.275 \pm 0.007$ & $0.5820 \pm 0.0108$ & $0.016 \pm 0.001$ & 34.6 / 75.0 & 43.7 / 24.2 \\
Shortcut $\times$ FIARSE & $0.1968 \pm 0.0018$ & $0.307 \pm 0.006$ & $0.5655 \pm 0.0074$ & $0.024 \pm 0.006$ & 76.1 / 93.7 & 75.3 / 93.5 \\
Shortcut $\times$ FedAvg & $0.1965 \pm 0.0023$ & $0.308 \pm 0.009$ & $0.5635 \pm 0.0063$ & $0.027 \pm 0.005$ & 76.1 / 93.7 & 75.3 / 93.5 \\
Width $\times$ FedAvg\textsuperscript{$\dagger$} & $0.1925 \pm 0.0057$ & $0.272 \pm 0.002$ & $0.5518 \pm 0.0123$ & $0.017 \pm 0.003$ & 74.9 / 93.6 & 74.9 / 93.7 \\
Width $\times$ FIARSE\textsuperscript{$\dagger$} & $0.1907 \pm 0.0011$ & $0.280 \pm 0.002$ & $0.5515 \pm 0.0019$ & $0.017 \pm 0.002$ & 74.9 / 93.6 & 74.9 / 93.7 \\
Width $\times$ FedRolex\textsuperscript{$\dagger$} & $0.1743 \pm 0.0157$ & $0.288 \pm 0.009$ & $0.5202 \pm 0.0281$ & $0.028 \pm 0.008$ & 74.9 / 93.6 & 74.9 / 93.7 \\
Depth $\times$ FjORD w/ KD & $0.1693 \pm 0.0101$ & $0.257 \pm 0.003$ & $0.5081 \pm 0.0143$ & $0.021 \pm 0.003$ & 96.9 / 99.2 & 74.4 / 93.8 \\
Depth $\times$ FjORD & $0.1648 \pm 0.0040$ & $0.253 \pm 0.002$ & $0.5053 \pm 0.0100$ & $0.015 \pm 0.001$ & 96.9 / 99.2 & 74.4 / 93.8 \\
Depth $\times$ FedAvg & $0.1607 \pm 0.0061$ & $0.281 \pm 0.009$ & $0.4988 \pm 0.0147$ & $0.027 \pm 0.005$ & 96.9 / 99.2 & 74.4 / 93.8 \\
\bottomrule
\end{tabular}
\end{table}

%% file: rq3-generality.tex
\begin{table}[t]
\centering
\footnotesize
\caption{Generality across continual-learning methods. Fleet-mean class-IL ACC (mean $\pm$ sample sd over $n{=}3$ seeds) for the three leading fleet geometries, each trained under four CL objectives on an identical replay buffer (6{,}400, uniform allocation) and an identical budget. Columns are reduction geometries and rows are objectives. DROD is highest in every row; bold marks that lead, and the EWC row carries none because all three fleets collapse there and the $0.004$ separating them is negligible. Per-tier values, forgetting and Task-IL are in Table~\ref{tab:rq3-clmethod}.}
\label{tab:rq3-generality}
\begin{tabular}{@{}lccc@{}}
\toprule
CL objective & DROD $\theta{=}0.50$ & Width $\times$ FjORD & Shortcut $\times$ FjORD \\
\midrule
ER & $\mathbf{0.3125} \pm 0.0160$ & $0.2741 \pm 0.0084$ & $0.2518 \pm 0.0143$ \\
DER++ & $\mathbf{0.2958} \pm 0.0008$ & $0.2745 \pm 0.0087$ & $0.2506 \pm 0.0098$ \\
BiC & $\mathbf{0.2308} \pm 0.0076$ & $0.2047 \pm 0.0115$ & $0.2003 \pm 0.0070$ \\
EWC & $0.0645 \pm 0.0010$ & $0.0609 \pm 0.0020$ & $0.0621 \pm 0.0007$ \\
\bottomrule
\end{tabular}
\end{table}

%% file: conclusion.tex
\section{Conclusion}
\label{sec:conclusion}

We studied federated continual learning when clients differ not only in their
data streams but also in the model capacity they can support. DROD extends
Ordered Dropout by separating the shortcut width from the residual-branch
width and sampling nested submodels along both dimensions. On our shared
budget ladder, DROD achieves the highest mean among the
compared heterogeneous FL strategies, benefits every capacity tier, and remains strongest
across the tested replay budgets and continual-learning objectives, except in
the EWC regime where all geometries collapse. These findings show that how a
residual network spends a fixed capacity budget matters: maintaining the
carried representation while compressing the residual transform is a
promising design principle for heterogeneous federated learning under drift.

%% file: appendix.tex
\appendix

\input{discussion}

\input{limitations}

\input{related-work}

\clearpage
\section{Hyperparameters and protocol details}
\label{sec:hparams}

Table~\ref{tab:hparams} lists every setting behind the results in
Section~\ref{sec:setup}. The recipe is matched line-by-line to Mammoth's
\texttt{seq-cifar100} configuration wherever a counterpart exists; the
remaining differences are the ones that exist \emph{because} this is
federated continual learning and Mammoth is centralized, and are marked as
such.

\begin{table}[htbp]
\centering
\footnotesize
\caption{Complete experimental configuration. ``FL'' marks a value that
differs from the centralized reference recipe because of the federated
structure itself, not as a tuning choice.}
\label{tab:hparams}
\begin{tabular}{@{}lll@{}}
\toprule
& setting & value \\
\midrule
\textit{Data}
 & dataset & CIFAR-100, 10 tasks $\times$ 10 classes \\
 & class order & fixed (identity), not shuffled across seeds \\
 & augmentation & random crop $32$, pad $4$ (black fill) $+$ horizontal flip \\
 & batch size & 32 (train), 512 (eval) \\
 & client partition & Dirichlet $\alpha{=}0.5$, redrawn per task \hfill \emph{FL} \\
\midrule
\textit{Model}
 & architecture & ResNet-18, CIFAR stem, from scratch \\
 & normalization & GroupNorm, 8 groups where divisible (App.~\ref{app:groupnorm}) \hfill \emph{FL} \\
 & full-capacity params & $11{,}220{,}132$ \\
 & full-capacity FLOPs & $1.111\times10^{9}$ (forward, batch 1) \\
\midrule
\textit{Optimization}
 & optimizer & SGD \\
 & learning rate & $0.1$, constant \\
 & momentum & $0.0$ \\
 & weight decay & $0.0$ \\
 & gradient clipping & none \\
 & local epochs / rounds & 4 epochs $\times$ 25 rounds per task \hfill \emph{FL} \\
\midrule
\textit{Federation}
 & clients & 10, full participation each round \\
 & tier assignment & round-robin, fixed for the run (4/3/3) \\
 & aggregation & per-index mean, weighted by client sample count \\
 & shared budget ladder & $1.000/0.251/0.064$ params; $1.000/0.251/0.063$ FLOPs \\
\midrule
\textit{Continual}
 & method & experience replay (reservoir) \\
 & buffer totals & 200 / 3{,}200 / 6{,}400 / 12{,}800 (20/320/640/1{,}280 per client) \\
 & per-tier allocation & uniform ($p{=}0$) unless stated \\
\midrule
\textit{Reporting}
 & seeds & 0, 1, 2 ($n{=}3$) \\
 & error bars & $\pm$ one sample standard deviation \\
\bottomrule
\end{tabular}
\end{table}

\paragraph{Choice of buffer totals.} The totals are of the form
$200 \cdot 2^k$. 200 is the smallest standard buffer of
DER++~\citep{buzzega2020derpp}, and 6{,}400 is the total closest to its
largest, 5{,}120; 3{,}200 and 12{,}800 bracket it by a factor of two.
DER++'s buffers are centralized. Federated class-incremental replay methods
give each client more on CIFAR-100: 1{,}000 exemplars in
Re-Fed~\citep{li2024refed} and 2{,}000 in GLFC~\citep{dong2022glfc}, against
our 640 (6{,}400 over ten clients).

\paragraph{Why GroupNorm.}
\label{app:groupnorm}
BatchNorm normalizes by minibatch statistics, so its estimates diverge across
clients holding different label distributions, and averaging the resulting
parameters is not a meaningful operation. GroupNorm normalizes per sample and
is therefore independent of a client's label distribution. The trade is
asymmetric in our favor: on CIFAR-10 under FedAvg, replacing BatchNorm with
GroupNorm costs $0.6$ points when clients are IID ($80.3\% \to 79.7\%$) and
recovers $19.8$ points when they are not ($50.2\% \to 70.0\%$), and the same
work shows the effect does not require extreme skew --- a partially non-IID
partition already costs $1.5$--$3.0$ points. (Their numbers come from a
BN-LeNet with a tuned group \emph{size} of 2; we fix the group count at 8 and did not
tune it. Where a layer's channel count is not divisible by 8, as in some
DROD tiers, the count is halved until it divides.) Our clients hold a Dirichlet($\alpha{=}0.5$) label
skew redrawn every task, so we take the IID-side cost and keep the non-IID
robustness. This is a scope choice, not a validated ablation
(Section~\ref{sec:limitations}).

\paragraph{What changes across seeds.} The three seeds (0, 1, 2) differ only
in the draws that are a function of the run seed; every other row of
Table~\ref{tab:hparams} is identical across seeds. Concretely, a different
seed redraws:
\begin{itemize}
\item \textbf{the per-task client partition} -- the Dirichlet split
  ($\alpha{=}0.5$) is redrawn independently for every task, so both which
  samples and how many samples each client holds change with the seed, task
  by task;
\item \textbf{local training order} -- minibatch order and augmentation for
  every (task, round, client) triple;
\item \textbf{Ordered-Dropout width sampling}, for the FjORD-style arms --
  which sub-model width each client draws at each local step;
\item \textbf{reservoir buffer contents} -- which exemplars survive eviction
  once a client's buffer is full;
\item \textbf{the initial weights}, via one global seed drawn at the start of
  the run.
\end{itemize}
Class order is \emph{not} on this list: every run in this paper sets
\texttt{shuffle\_classes=false} (Table~\ref{tab:hparams}), so all three seeds
see the identical class-to-task assignment; the Dirichlet redraw above
changes each client's per-task sample composition, not the task boundaries
themselves. The learning rate, architecture, budget ladder, buffer totals,
and tier assignment are likewise fixed hyperparameters, not seed draws.

\paragraph{What the budget match does and does not fix.} The shared ladder
fixes \emph{parameters} and \emph{inference FLOPs}, and those two only. Three
further cost axes move freely.

\begin{itemize}
\item \textbf{Peak activation memory is not matched, by construction.} On the
  DROD axis the peak intermediate tensor is the stem output, which the branch
  ratio never touches, so activation memory is governed by the shortcut ratio
  alone. An interior $\theta$ therefore \emph{exceeds} a width-sliced tier of
  the same parameter budget on activation memory, up to the full-width value
  at $\theta{=}1$. This is a genuine cost of the method, not a measurement
  artifact as DROD buys its accuracy at equal parameters and equal inference
  FLOPs and pays for it in activation footprint.
\item \textbf{Training FLOPs are not matched either, and this is inherited
  from Ordered Dropout rather than introduced by DROD.} Sampling a sub-model
  per iteration means the realized per-iteration training cost of every
  Ordered-Dropout arm sits above its own parameter/inference-FLOP label; we
  measure it separately rather than folding it into the match, and it is
  common to DROD and to the FjORD baselines it is compared against.
\item \textbf{Replay enters the local batch by concatenation.} For every arm
  trained under Ordered Dropout, buffer samples are appended to the batch
  before the forward pass rather than contributing a separately weighted loss
  term. Once the buffer is populated this means such a step performs a
  forward/backward pass over up to twice as many samples as a replay-free
  step. Step counts are matched across arms; per-step compute is not.
\end{itemize}

\paragraph{A consequence of splitting the buffer across clients.} At the
200-exemplar total, each client holds 20 exemplars while the replay batch is
32, so every local step replays that client's \emph{entire} buffer rather than
a random draw from it. Splitting one fleet-wide exemplar budget across ten
clients necessarily lowers per-step replay diversity relative to a centralized
buffer of the same total size, and no setting of the replay batch recovers it
without un-matching the budget. We disclose this rather than compensating for
it, since it is a property of federated replay at a matched budget and applies
identically to every arm compared here.

\clearpage
\section{Full tabular results}
\label{app:full-results}

\paragraph{Why the grid is not a full $m \times n$.}
\label{app:grid-shape}
A few aggregation rules are only defined for one reduction axis.
FedRolex~\citep{alam2022fedrolex} slides a fixed-width window across
channel indices, relying on those indices keeping a stable identity from
round to round so that every channel is eventually trained by every tier;
neither which residual blocks survive (depth) nor the branch's hidden rank
(shortcut) has such an identity to slide over, so we implement FedRolex on
the width axis only, giving a single Width~$\times$~FedRolex row rather
than one per axis. ScaleFL~\citep{ilhan2023scalefl} is missing from the
grid for a related reason: it fixes its own combined width/depth split per
tier rather than taking a reduction axis as an external input, so it
appears once, on its own terms, rather than crossed with any axis.

\input{rq3-grid-full}

\input{rq3-clmethod-full}

\input{rq3-bufp}

\input{rq3-bufp-full}

\FloatBarrier

\section{Additional method illustration}
\label{app:method-illustration}

\begin{figure}[h]
\centering
\includegraphics[width=0.86\linewidth]{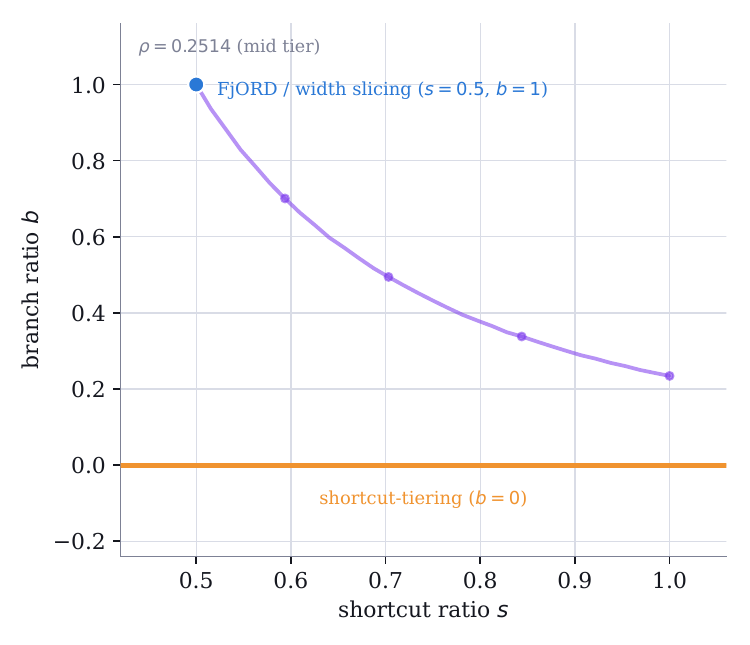}
\caption{The $(s,b)$ iso-budget curve $bs^2=\rho$ for our mid tier
($\rho{=}0.2514$), with the two literature baselines marked
explicitly. FjORD / width slicing sits at $\theta{=}0$, the single point
$(s{=}0.5,\,b{=}1)$ where the shortcut and branch narrow together;
shortcut-tiering's $b\to0$ precursor is drawn as the $b{=}0$ axis itself,
reachable at any $s$ but off this particular curve, since it is a
discrete, per-block toggle rather than a point on a nonzero-$\rho$ curve.
DROD spans the full nonzero-budget curve from the FjORD endpoint at
$\theta{=}0$ to the full-shortcut, low-rank endpoint at $\theta{=}1$;
shortcut-tiering is shown separately as the limiting $b{=}0$ axis.}
\label{fig:theta-curve}
\end{figure}

%% file: discussion.tex
\section{Discussion}
\label{sec:discussion}

Our results challenge the usual identification of a client's capacity with a
single width multiplier.  A residual block contains two qualitatively
different resources: the shortcut carries the representation forward, while
the residual branch transforms it.  Narrowing both by the same factor saves
compute, but also constricts the representation available to every subsequent
block.  DROD can instead preserve a wider carried representation and spend the
remaining budget on a lower-rank transform.  The fact that an interior split
($\theta{=}0.50$), rather than either endpoint, has the highest mean suggests
that neither resource should be preserved at the complete expense of the
other.  A full-width shortcut with too narrow a branch limits how much new
structure each block can learn, whereas ordinary width slicing gives the
branch more hidden channels only by narrowing the stream on which it operates.
Thus, the relevant design question is not simply how large a submodel is, but
where in a residual block its limited capacity is placed.

This allocation also changes which parameters receive useful federated
updates.  Under prefix-sliced aggregation, low-index coordinates are shared by
more client tiers and are consequently trained more often.  With one width
ratio, reducing a tier simultaneously removes coordinates from the carried
stream and from the branch.  DROD separates these effects: smaller clients can
continue to update a comparatively broad shared stream while contributing a
narrower residual transform.  This provides a plausible explanation for why
the benefit is not confined to the small models.  All three tiers improve,
including the full-capacity tier whose own inference architecture is not
compressed, indicating that the geometry changes the quality of the shared
training signal rather than merely producing a better isolated small model.
The particularly weak performance of fixed width slices and FedRolex, relative
to the Ordered-Dropout variants, further suggests that allocating coordinates
well and exposing nested submodels repeatedly during local training are
complementary rather than interchangeable choices.

The continual-learning experiments help delimit this interpretation.  DROD
has the highest mean with ER, DER++, and BiC, and it remains ahead as the replay
buffer changes, so its advantage is unlikely to be an artifact of one replay
loss or one memory size.  At the same time, the absolute accuracy of every
geometry depends strongly on the continual-learning mechanism: under EWC all
three collapse to about 0.06 class-IL accuracy, and the differences between them
are negligible.  Reduction geometry therefore cannot compensate for an
objective that fails to retain old classes in this single-head regime.  DROD
should be understood as complementary to continual learning: it determines
how heterogeneous clients represent and aggregate what they learn, while the
local objective still determines whether past knowledge survives at all.

There is also a systems tradeoff behind the accuracy gain.  Matching parameters
and inference FLOPs does not make two submodels equally deployable.  At a fixed
budget, DROD's wider shortcut produces wider activations than ordinary width
slicing, increasing peak activation memory; Ordered Dropout also makes realized
training cost depend on the submodels sampled during local optimization.
Consequently, $\theta$ is best viewed as a hardware-aware allocation knob, not
as a universally optimal constant.  Devices constrained mainly by model
storage, communication, or inference compute may benefit from moving capacity
toward the shortcut, whereas devices constrained by working memory may prefer
a smaller $\theta$.  Measuring this Pareto frontier on physical devices would
be more informative than comparing parameter counts alone.

More broadly, the study points to residual structure as an underused source of
flexibility in heterogeneous learning.  Existing schemes largely choose among
width, depth, or unstructured masks; the DROD result shows that even within a
single block, input/output width and transformation rank need not be coupled.
The same principle may extend to bottleneck residual networks and transformer
feed-forward blocks, although the best allocation is likely to depend on the
architecture, normalization, and task stream.  The present evidence supports
the narrower claim that, for the tested ResNet-18 FCL setting, jointly ordering
the shortcut and branch is a more effective use of a matched budget than
ordering either structural axis alone.

%% file: limitations.tex
\section{Limitations}
\label{sec:limitations}

Our evidence is limited to Split-CIFAR-100, one ResNet-18 backbone, ten
clients, synchronous full participation, and three seeds. The comparison
therefore does not establish that the same ordering holds for larger
datasets, non-residual architectures, partial participation, asynchronous
training, or more severe system and data heterogeneity. Although parameters
and inference FLOPs are matched, peak activation memory is not; retaining a wider shortcut can require more activation memory
than width slicing at the same nominal budget. Normalization is a scope choice
we did not ablate. Every arm uses GroupNorm, adopted on
\citeauthor{hsieh2020quagmire}'s evidence rather than on a
GroupNorm-vs-BatchNorm comparison in our own setting, and they themselves note
that more study is needed before GroupNorm can be assumed to replace BatchNorm
across models and applications. Extending the comparison to BatchNorm is the
natural next step. Most of the grid
also uses experience replay with uniformly allocated local buffers. Finally, DROD's lead over its closest rival, Width~$\times$~FjORD, depends on
the buffer: it holds on all three seeds at our main buffer of 6{,}400 but
narrows to $0.021$ and two of three seeds at 12{,}800. Larger-scale experiments,
additional seeds, and measurements on physical heterogeneous devices are
needed to determine the robustness and practical cost of the observed gains.

%% file: related-work.tex
\section{Related work}
\label{sec:related-work}

\paragraph{Model-heterogeneous federated learning.}
Partial-training methods let clients of unequal capacity train subsets of a
single global model and average each parameter over exactly those clients whose
submodel contained it. Federated Dropout~\citep{caldas2018feddropout}
introduced the idea with randomly selected subnetworks; subsequent work makes
the selection structured, so that a smaller client's model nests inside a larger
one. HeteroFL~\citep{diao2021heterofl} assigns each client a fixed contiguous
band of channels in every layer. FedRolex~\citep{alam2022fedrolex} rotates that
band across channel indices between rounds, so that every parameter is
eventually trained by every tier and no region of the global model is updated
only by the largest clients. FjORD~\citep{horvath2021fjord} instead resamples a
width rate at each local iteration, capped by the client's device ceiling:
low-index channels are exercised by every draw and high-index channels only by
the widest, which induces an importance ordering on channels ---
\emph{Ordered Dropout} --- and wider submodels are distilled into narrower
ones. Slimmable and universally slimmable
networks~\citep{yu2019slimmable,yu2019usnets} develop the same
width-switchable training centrally, and
Split-Mix~\citep{hong2022splitmix} trains a set of small base subnetworks that
are recombined on demand at inference time.

Depth is the natural alternative axis. DepthFL~\citep{kim2023depthfl} prunes the
deepest layers, attaches a classifier at each retained depth, and mutually
distils among them, reporting that depth-scaled local models build a better
global model than width-scaled ones on stationary CIFAR-100.
FeDepth~\citep{fedepth2023} decomposes the network into blocks trained
sequentially to fit a client's memory budget, targeting activation memory rather
than the aggregation rule. ScaleFL~\citep{ilhan2023scalefl} divides the budget
along width and depth simultaneously, holding the two ratios equal, and ties
early exits to the final classifier through self-distillation.
InclusiveFL~\citep{liu2022inclusivefl} allocates transformers of differing depth
and transfers knowledge from powerful to weak clients by momentum distillation.
FIARSE~\citep{wu2024fiarse} abandons fixed structural slices altogether,
selecting each client's submodel by parameter magnitude and updating that choice
as training proceeds; because the resulting mask is unstructured, it reduces
parameter and communication cost but not computation, since a dense kernel does
identical work on a masked tensor. Throughout this literature, evaluation is at
a single stationary data distribution.

\paragraph{Federated continual learning.}
Work on clients whose data arrive as a non-stationary stream has largely assumed
one common architecture. FedWeIT~\citep{yoon2021fedweit} decomposes each
client's parameters into a dense shared base and sparse task-adaptive
components, so that knowledge transfers selectively between clients rather than
through a single average. FedTA~\citep{fedta2025} mixes trainable tail anchors
with the frozen features of a pretrained backbone to counter forgetting both
between clients and between tasks, and DOLFIN~\citep{moussadek2025dolfin}
equips a shared vision transformer with low-rank adapters constrained by dual
gradient projection memory. STAMP~\citep{stamp2025}, notwithstanding
``heterogeneous'' in its title, addresses task and data heterogeneity: clients
draw unrelated or even conflicting tasks from a large pool, while every client
trains the same network. In all of these the capacity available to a client is
uniform, and heterogeneity resides in the data.

\paragraph{Heterogeneous architectures under drift.}
Few works vary architecture and data stream together, and those that do share
a design. FedDCL~\citep{feddcl2025} trains ResNet-18/32/50 clients that never
exchange weights, sending synthetic prototypes drawn from a frozen pretrained
diffusion model to the server instead; on its CIFAR-100 benchmark --- 30
classes over three tasks, five clients --- the server is a pretrained vision
transformer of which only the classifier head is fine-tuned, and its
from-scratch configuration is confined to shallow convolutional networks on
grayscale MNIST-family data. Fed-LSCL~\citep{fedlscl2025} pairs five different
CNN families with one shared frozen transformer through a per-client linear
generator that emits a LoRA adapter, fused on the server by distillation over
buffered features, across five tasks of eight classes and likewise five
clients. Both freeze a shared pretrained model and exchange only distilled or
generated signal, which removes the aggregation-shape problem rather than
solving it and leaves the representation-bearing model untrained under
federation. Both are also evaluated on short streams covering a fraction of
the label space. SpaPFCIL~\citep{liu2025spapficl} is closer in spirit,
learning a personalized sparse class-incremental model per client, but every
client shares one backbone and one global sparsity target and differs only in
which weights its mask selects, so capacity is uniform by construction. We
instead train a ResNet-18 fleet from scratch, with clients of genuinely
different capacity contributing into a single shared weight space, over ten
tasks spanning all one hundred classes, with ten clients and no pretrained
model anywhere in the system.

\paragraph{Continual-learning objectives.}
We hold the federated geometry fixed and vary the objective clients optimise
locally, drawing on the standard families. Regularisation methods constrain
movement in parameter space: EWC~\citep{kirkpatrick2017ewc} penalises deviation
from the previous task's solution in proportion to the diagonal Fisher
information, and LwF~\citep{li2016lwf} distils the previous model's outputs on
new-task data without storing any examples. Rehearsal methods keep a bounded
exemplar memory: experience replay~\citep{chaudhry2019er} interleaves sampled
exemplars with the incoming batch, and iCaRL~\citep{rebuffi2017icarl} combines
herded exemplars with a nearest-mean classifier.
DER++~\citep{buzzega2020derpp} strengthens replay by also matching the logits
recorded when an exemplar was stored, and BiC~\citep{wu2019bic} targets the
classifier's bias toward recently seen classes directly, correcting it with a
two-parameter linear rescaling fitted on a balanced held-out split. We report
the class-incremental scenario throughout~\citep{vandeven2019three}, in which no
task identifier is available at test time. GDumb~\citep{prabhu2020gdumb} is a
standing caution for this family: a learner that merely stores a class-balanced
memory and retrains from scratch at test time is competitive with much of it, so
the replay budget must be held fixed across compared systems, as it is here.

\paragraph{Ordered representations and residual scaling.}
DROD composes two lines of work that have not previously been brought together
under federated aggregation with drift. The first is ordered, nested subnetwork
sampling. Nested dropout~\citep{rippel2014nesteddropout} stochastically removes
coherent nested sets of units, which orders the representation by importance and,
for a semi-linear autoencoder, recovers PCA exactly; Matryoshka representation
learning~\citep{kusupati2022matryoshka} and MatFormer~\citep{devvrit2024matformer}
carry the same coarse-to-fine nesting into embeddings and into transformer
feed-forward blocks, and FjORD's Ordered Dropout is its federated instance. The
second is the treatment of a residual block as an interpolation between the
identity and a full transform. Fixup, SkipInit, ReZero and
LayerScale~\citep{zhang2019fixup,de2020skipinit,bachlechner2021rezero,touvron2021layerscale}
each bias a block toward the identity by attenuating its residual branch with a
small or zero-initialised scale, while stochastic depth and
LayerDrop~\citep{huang2016stochasticdepth,fan2020layerdrop} drop whole blocks at
random, consistent with the view of a residual network as an ensemble of paths
of varying length~\citep{veit2016resnetensembles}. DROD performs this
interpolation structurally rather than multiplicatively: the branch ratio
narrows the residual branch's hidden width instead of scaling its output, so
approaching the identity releases parameters and FLOPs rather than leaving them
in place, and the shortcut ratio independently sets the width of the carried
stream.

%% file: rq3-grid-full.tex
\begin{table}[htbp]
\centering
\footnotesize
\caption{Per-tier class-IL ACC on the shared \emph{parameter} ladder, buf=6{,}400, $n{=}3$, mean $\pm$ sample sd, sorted by fleet mean. The high/mid/low columns give the per-tier split behind each fleet mean (tier keys t100/t025/t006); the fleet column is their unweighted mean. This is the params-matched companion to the FLOPs-matched main grid (Table~\ref{tab:rq3-main}), which reports the same fleet means without the per-tier breakdown. The DROD rows span the $\theta$ axis: $\theta{=}1.00$ is the full-width-shortcut endpoint, $\theta{\in}\{0.25,0.50,0.75\}$ interior; the width-slicing endpoint $\theta{=}0$ is the Width~$\times$~FjORD (kd=false) row itself. $^{\dagger}$sits on both ladders -- narrowing channels cuts parameters and inference FLOPs together, so one run meets both budgets. $^{\ddagger}$ScaleFL keeps its own combined width/depth split and sits on neither ladder.}
\vspace{1em}
\label{tab:rq3-grid-params-6400}
\begin{tabular}{@{}lcccc@{}}
\toprule
cell & fleet class-IL ACC & high (t100) & mid (t025) & low (t006) \\
\midrule
DROD $\theta$=0.50$^\dagger$ & 0.3125 $\pm$ 0.0160 & 0.3076 & 0.3161 & 0.3137 \\
DROD $\theta$=1.00$^\dagger$ & 0.2812 $\pm$ 0.0048 & 0.2795 & 0.2840 & 0.2801 \\
DROD $\theta$=0.25$^\dagger$ & 0.2760 $\pm$ 0.0091 & 0.2709 & 0.2785 & 0.2784 \\
Width $\times$ FjORD (kd=false)$^\dagger$ & 0.2741 $\pm$ 0.0084 & 0.2701 & 0.2815 & 0.2708 \\
DROD $\theta$=0.75$^\dagger$ & 0.2723 $\pm$ 0.0083 & 0.2680 & 0.2766 & 0.2723 \\
Width $\times$ FjORD (kd=true)$^\dagger$ & 0.2719 $\pm$ 0.0157 & 0.2493 & 0.2896 & 0.2770 \\
Shortcut $\times$ FjORD (kd=false) & 0.2486 $\pm$ 0.0153 & 0.2498 & 0.2482 & 0.2479 \\
Shortcut $\times$ FIARSE & 0.2474 $\pm$ 0.0067 & 0.2479 & 0.2595 & 0.2349 \\
Shortcut $\times$ FjORD (kd=true) & 0.2414 $\pm$ 0.0045 & 0.2398 & 0.2389 & 0.2454 \\
Depth $\times$ FIARSE & 0.2187 $\pm$ 0.0070 & 0.2417 & 0.2287 & 0.1858 \\
ScaleFL$^\ddagger$ & 0.2149 $\pm$ 0.0054 & 0.2405 & 0.1934 & 0.2110 \\
Depth $\times$ FjORD (kd=false) & 0.2088 $\pm$ 0.0037 & 0.2367 & 0.2042 & 0.1855 \\
Depth $\times$ FjORD (kd=true) & 0.2076 $\pm$ 0.0080 & 0.2338 & 0.2032 & 0.1857 \\
Width $\times$ FedAvg$^\dagger$ & 0.1925 $\pm$ 0.0057 & 0.2308 & 0.1905 & 0.1562 \\
Width $\times$ FIARSE$^\dagger$ & 0.1907 $\pm$ 0.0011 & 0.2258 & 0.1899 & 0.1563 \\
Shortcut $\times$ FedAvg & 0.1888 $\pm$ 0.0190 & 0.2026 & 0.1922 & 0.1715 \\
Depth $\times$ FedAvg & 0.1882 $\pm$ 0.0092 & 0.2241 & 0.1745 & 0.1660 \\
Width $\times$ FedRolex$^\dagger$ & 0.1743 $\pm$ 0.0157 & 0.2557 & 0.1860 & 0.0812 \\
\bottomrule
\end{tabular}
\end{table}

\begin{table}[p]
\centering
\footnotesize
\caption{Same grid on the shared \emph{inference-FLOP} ladder, buf=6{,}400. The $\dagger$ rows (width-family and DROD) are the same runs as on the parameter ladder -- parameters and FLOPs move together for those axes (Section~\ref{sec:method}) -- and are repeated here so the ranking is complete; the depth and shortcut rows are re-solved to hit the FLOP target and so differ from Table~\ref{tab:rq3-grid-params-6400}. Depth~$\times$~FIARSE has no cell on this ladder (a magnitude mask removes no multiplies; see Table~\ref{tab:rq3-main}). $^{\ddagger}$ScaleFL sits on neither ladder.}
\vspace{1em}
\label{tab:rq3-grid-flops-6400}
\begin{tabular}{@{}lcccc@{}}
\toprule
cell & fleet class-IL ACC & high (t100) & mid (t025) & low (t006) \\
\midrule
DROD $\theta$=0.50$^\dagger$ & 0.3125 $\pm$ 0.0160 & 0.3076 & 0.3161 & 0.3137 \\
DROD $\theta$=1.00$^\dagger$ & 0.2812 $\pm$ 0.0048 & 0.2795 & 0.2840 & 0.2801 \\
DROD $\theta$=0.25$^\dagger$ & 0.2760 $\pm$ 0.0091 & 0.2709 & 0.2785 & 0.2784 \\
Width $\times$ FjORD (kd=false)$^\dagger$ & 0.2741 $\pm$ 0.0084 & 0.2701 & 0.2815 & 0.2708 \\
DROD $\theta$=0.75$^\dagger$ & 0.2723 $\pm$ 0.0083 & 0.2680 & 0.2766 & 0.2723 \\
Width $\times$ FjORD (kd=true)$^\dagger$ & 0.2719 $\pm$ 0.0157 & 0.2493 & 0.2896 & 0.2770 \\
Shortcut $\times$ FjORD (kd=false) & 0.2518 $\pm$ 0.0143 & 0.2530 & 0.2459 & 0.2564 \\
Shortcut $\times$ FjORD (kd=true) & 0.2368 $\pm$ 0.0097 & 0.2361 & 0.2256 & 0.2485 \\
ScaleFL$^\ddagger$ & 0.2149 $\pm$ 0.0054 & 0.2405 & 0.1934 & 0.2110 \\
Shortcut $\times$ FIARSE & 0.1968 $\pm$ 0.0018 & 0.2231 & 0.1921 & 0.1754 \\
Shortcut $\times$ FedAvg & 0.1965 $\pm$ 0.0023 & 0.2240 & 0.1864 & 0.1793 \\
Width $\times$ FedAvg$^\dagger$ & 0.1925 $\pm$ 0.0057 & 0.2308 & 0.1905 & 0.1562 \\
Width $\times$ FIARSE$^\dagger$ & 0.1907 $\pm$ 0.0011 & 0.2258 & 0.1899 & 0.1563 \\
Width $\times$ FedRolex$^\dagger$ & 0.1743 $\pm$ 0.0157 & 0.2557 & 0.1860 & 0.0812 \\
Depth $\times$ FjORD (kd=true) & 0.1693 $\pm$ 0.0101 & 0.2207 & 0.1434 & 0.1437 \\
Depth $\times$ FjORD (kd=false) & 0.1648 $\pm$ 0.0040 & 0.2110 & 0.1359 & 0.1474 \\
Depth $\times$ FedAvg & 0.1607 $\pm$ 0.0061 & 0.2154 & 0.1373 & 0.1294 \\
\bottomrule
\end{tabular}
\end{table}

\begin{table}[p]
\centering
\footnotesize
\caption{Same grid, \emph{parameter} ladder, buf=200. Columns, sorting and the $\dagger$/$\ddagger$ markers as in Table~\ref{tab:rq3-grid-params-6400}.}
\label{tab:rq3-grid-params-200}
\begin{tabular}{@{}lcccc@{}}
\toprule
cell & fleet class-IL ACC & high (t100) & mid (t025) & low (t006) \\
\midrule
DROD $\theta$=0.50$^\dagger$ & 0.0863 $\pm$ 0.0044 & 0.0832 & 0.0870 & 0.0886 \\
DROD $\theta$=0.25$^\dagger$ & 0.0845 $\pm$ 0.0032 & 0.0828 & 0.0848 & 0.0861 \\
DROD $\theta$=1.00$^\dagger$ & 0.0845 $\pm$ 0.0002 & 0.0833 & 0.0849 & 0.0854 \\
DROD $\theta$=0.75$^\dagger$ & 0.0822 $\pm$ 0.0029 & 0.0806 & 0.0832 & 0.0829 \\
Width $\times$ FjORD (kd=true)$^\dagger$ & 0.0818 $\pm$ 0.0036 & 0.0790 & 0.0819 & 0.0845 \\
Shortcut $\times$ FIARSE & 0.0798 $\pm$ 0.0016 & 0.0804 & 0.0810 & 0.0779 \\
Width $\times$ FjORD (kd=false)$^\dagger$ & 0.0797 $\pm$ 0.0001 & 0.0778 & 0.0794 & 0.0819 \\
Shortcut $\times$ FjORD (kd=false) & 0.0780 $\pm$ 0.0009 & 0.0767 & 0.0781 & 0.0792 \\
Shortcut $\times$ FjORD (kd=true) & 0.0766 $\pm$ 0.0044 & 0.0761 & 0.0773 & 0.0765 \\
Depth $\times$ FIARSE & 0.0744 $\pm$ 0.0063 & 0.0818 & 0.0790 & 0.0623 \\
Depth $\times$ FjORD (kd=true) & 0.0714 $\pm$ 0.0040 & 0.0769 & 0.0712 & 0.0659 \\
Depth $\times$ FjORD (kd=false) & 0.0708 $\pm$ 0.0063 & 0.0780 & 0.0706 & 0.0640 \\
ScaleFL$^\ddagger$ & 0.0701 $\pm$ 0.0051 & 0.0763 & 0.0668 & 0.0673 \\
Shortcut $\times$ FedAvg & 0.0690 $\pm$ 0.0054 & 0.0753 & 0.0693 & 0.0625 \\
Width $\times$ FIARSE$^\dagger$ & 0.0660 $\pm$ 0.0063 & 0.0758 & 0.0672 & 0.0552 \\
Depth $\times$ FedAvg & 0.0656 $\pm$ 0.0039 & 0.0748 & 0.0626 & 0.0594 \\
Width $\times$ FedAvg$^\dagger$ & 0.0637 $\pm$ 0.0065 & 0.0738 & 0.0629 & 0.0545 \\
Width $\times$ FedRolex$^\dagger$ & 0.0589 $\pm$ 0.0043 & 0.0776 & 0.0597 & 0.0393 \\
\bottomrule
\end{tabular}
\end{table}

\begin{table}[p]
\centering
\footnotesize
\caption{Same grid, \emph{inference-FLOP} ladder, buf=200. Columns, sorting and the $\dagger$/$\ddagger$ markers as in Table~\ref{tab:rq3-grid-params-6400}; Depth~$\times$~FIARSE again has no cell on this ladder.}
\label{tab:rq3-grid-flops-200}
\begin{tabular}{@{}lcccc@{}}
\toprule
cell & fleet class-IL ACC & high (t100) & mid (t025) & low (t006) \\
\midrule
DROD $\theta$=0.50$^\dagger$ & 0.0863 $\pm$ 0.0044 & 0.0832 & 0.0870 & 0.0886 \\
DROD $\theta$=0.25$^\dagger$ & 0.0845 $\pm$ 0.0032 & 0.0828 & 0.0848 & 0.0861 \\
DROD $\theta$=1.00$^\dagger$ & 0.0845 $\pm$ 0.0002 & 0.0833 & 0.0849 & 0.0854 \\
DROD $\theta$=0.75$^\dagger$ & 0.0822 $\pm$ 0.0029 & 0.0806 & 0.0832 & 0.0829 \\
Width $\times$ FjORD (kd=true)$^\dagger$ & 0.0818 $\pm$ 0.0036 & 0.0790 & 0.0819 & 0.0845 \\
Width $\times$ FjORD (kd=false)$^\dagger$ & 0.0797 $\pm$ 0.0001 & 0.0778 & 0.0794 & 0.0819 \\
Shortcut $\times$ FjORD (kd=false) & 0.0771 $\pm$ 0.0021 & 0.0766 & 0.0766 & 0.0782 \\
Shortcut $\times$ FjORD (kd=true) & 0.0757 $\pm$ 0.0043 & 0.0774 & 0.0726 & 0.0773 \\
ScaleFL$^\ddagger$ & 0.0701 $\pm$ 0.0051 & 0.0763 & 0.0668 & 0.0673 \\
Shortcut $\times$ FIARSE & 0.0700 $\pm$ 0.0039 & 0.0767 & 0.0709 & 0.0623 \\
Shortcut $\times$ FedAvg & 0.0673 $\pm$ 0.0059 & 0.0743 & 0.0654 & 0.0623 \\
Width $\times$ FIARSE$^\dagger$ & 0.0660 $\pm$ 0.0063 & 0.0758 & 0.0672 & 0.0552 \\
Width $\times$ FedAvg$^\dagger$ & 0.0637 $\pm$ 0.0065 & 0.0738 & 0.0629 & 0.0545 \\
Depth $\times$ FjORD (kd=false) & 0.0631 $\pm$ 0.0043 & 0.0750 & 0.0554 & 0.0587 \\
Depth $\times$ FjORD (kd=true) & 0.0615 $\pm$ 0.0059 & 0.0745 & 0.0542 & 0.0558 \\
Width $\times$ FedRolex$^\dagger$ & 0.0589 $\pm$ 0.0043 & 0.0776 & 0.0597 & 0.0393 \\
Depth $\times$ FedAvg & 0.0584 $\pm$ 0.0050 & 0.0732 & 0.0521 & 0.0499 \\
\bottomrule
\end{tabular}
\end{table}

%% file: rq3-clmethod-full.tex
\begin{table}[p]
\centering
\footnotesize
\caption{Generality across CL methods, full per-tier breakdown (companion to Table~\ref{tab:rq3-generality}). Buffer 6{,}400, uniform allocation, $n{=}3$, mean $\pm$ sample sd. Tiers are the shared ladder's high, mid and low rungs. In every one of the twelve method$\times$tier comparisons below, DROD's tier is above the corresponding tier of both baselines.}
\vspace{1em}
\label{tab:rq3-clmethod}
\setlength{\tabcolsep}{4pt}
\begin{tabular}{@{}llccccc@{}}
\toprule
CL objective & fleet & class-IL ACC & high & mid & low & class-IL forg. \\
\midrule
ER & \textbf{DROD} & $0.3125 \pm 0.0160$ & 0.3076 & 0.3161 & 0.3137 & 0.251 \\
 & Width & $0.2741 \pm 0.0084$ & 0.2701 & 0.2815 & 0.2708 & 0.263 \\
 & Shortcut & $0.2518 \pm 0.0143$ & 0.2530 & 0.2459 & 0.2564 & 0.291 \\
\addlinespace
DER++ & \textbf{DROD} & $0.2958 \pm 0.0008$ & 0.2947 & 0.2975 & 0.2952 & 0.322 \\
 & Width & $0.2745 \pm 0.0087$ & 0.2805 & 0.2835 & 0.2596 & 0.326 \\
 & Shortcut & $0.2506 \pm 0.0098$ & 0.2460 & 0.2501 & 0.2555 & 0.324 \\
\addlinespace
BiC & \textbf{DROD} & $0.2308 \pm 0.0076$ & 0.2342 & 0.2307 & 0.2274 & 0.226 \\
 & Width & $0.2047 \pm 0.0115$ & 0.2129 & 0.2108 & 0.1903 & 0.225 \\
 & Shortcut & $0.2003 \pm 0.0070$ & 0.2106 & 0.1985 & 0.1918 & 0.221 \\
\addlinespace
EWC & \textbf{DROD} & $0.0645 \pm 0.0010$ & 0.0631 & 0.0645 & 0.0660 & 0.532 \\
 & Width & $0.0609 \pm 0.0020$ & 0.0604 & 0.0613 & 0.0609 & 0.503 \\
 & Shortcut & $0.0621 \pm 0.0007$ & 0.0615 & 0.0619 & 0.0628 & 0.524 \\
\bottomrule
\end{tabular}
\end{table}

%% file: rq3-bufp.tex
\begin{table}[t]
\centering
\footnotesize
\caption{Fleet-mean class-IL ACC against replay-buffer total, uniform per-tier allocation (every tier gets the same buffer). Plain ER, $n{=}3$, mean $\pm$ sample sd. DROD is highest at all three buffer totals. Per-tier values: Table~\ref{tab:rq3-bufp-full}.}
\label{tab:rq3-bufp}
\setlength{\tabcolsep}{4pt}
\begin{tabular}{@{}lccc@{}}
\toprule
buffer & \textbf{DROD} $\theta{=}0.50$ & Width $\times$ FjORD & Shortcut $\times$ FjORD \\
\midrule
3,200 & $0.2414 \pm 0.0102$ & $0.2140 \pm 0.0094$ & $0.1918 \pm 0.0051$ \\
6,400 & $0.3125 \pm 0.0160$ & $0.2741 \pm 0.0084$ & $0.2518 \pm 0.0143$ \\
12,800 & $0.3547 \pm 0.0186$ & $0.3333 \pm 0.0204$ & $0.3025 \pm 0.0087$ \\
\bottomrule
\end{tabular}
\end{table}

%% file: rq3-bufp-full.tex
\begin{table}[p]
\centering
\footnotesize
\caption{Fleet-mean class-IL ACC against replay-buffer total, uniform per-tier allocation (every tier gets the same buffer). Plain ER, $n{=}3$, mean $\pm$ sample sd. Per-tier breakdown, one block per fleet; the fleet column repeats Table~\ref{tab:rq3-bufp}. DROD's high, mid and low tier are each above the corresponding tier of both baselines at all three buffer totals.}
\vspace{1em}
\label{tab:rq3-bufp-full}
\setlength{\tabcolsep}{4pt}
\begin{tabular}{@{}llcccc@{}}
\toprule
fleet & buffer & fleet & high & mid & low \\
\midrule
\textbf{DROD} $\theta{=}0.50$ & 3,200 & $0.2414 \pm 0.0102$ & 0.2341 & 0.2423 & 0.2478 \\
 & 6,400 & $0.3125 \pm 0.0160$ & 0.3076 & 0.3161 & 0.3137 \\
 & 12,800 & $0.3547 \pm 0.0186$ & 0.3516 & 0.3603 & 0.3523 \\
\addlinespace
Width $\times$ FjORD & 3,200 & $0.2140 \pm 0.0094$ & 0.2080 & 0.2170 & 0.2170 \\
 & 6,400 & $0.2741 \pm 0.0084$ & 0.2701 & 0.2815 & 0.2708 \\
 & 12,800 & $0.3333 \pm 0.0204$ & 0.3354 & 0.3472 & 0.3175 \\
\addlinespace
Shortcut $\times$ FjORD & 3,200 & $0.1918 \pm 0.0051$ & 0.1919 & 0.1879 & 0.1955 \\
 & 6,400 & $0.2518 \pm 0.0143$ & 0.2530 & 0.2459 & 0.2564 \\
 & 12,800 & $0.3025 \pm 0.0087$ & 0.3071 & 0.2966 & 0.3038 \\
\bottomrule
\end{tabular}
\end{table}

%% file: main.bbl
\begin{thebibliography}{42}
\providecommand{\natexlab}[1]{#1}
\providecommand{\url}[1]{\texttt{#1}}
\expandafter\ifx\csname urlstyle\endcsname\relax
  \providecommand{\doi}[1]{doi: #1}\else
  \providecommand{\doi}{doi: \begingroup \urlstyle{rm}\Url}\fi

\bibitem[Alam et~al.(2022)Alam, Liu, Yan, and Zhang]{alam2022fedrolex}
Samiul Alam, Luyang Liu, Ming Yan, and Mi~Zhang.
\newblock {F}ed{R}olex: Model-heterogeneous federated learning with rolling
  sub-model extraction.
\newblock In \emph{Advances in Neural Information Processing Systems
  (NeurIPS)}, 2022.

\bibitem[Bachlechner et~al.(2021)Bachlechner, Majumder, Mao, Cottrell, and
  McAuley]{bachlechner2021rezero}
Thomas Bachlechner, Bodhisattwa~Prasad Majumder, Huanru~Henry Mao, Garrison~W.
  Cottrell, and Julian McAuley.
\newblock {ReZero} is all you need: Fast convergence at large depth.
\newblock In \emph{Proceedings of the 37th Conference on Uncertainty in
  Artificial Intelligence}, volume 161 of \emph{Proceedings of Machine Learning
  Research}, pages 1352--1361, 2021.
\newblock URL \url{https://proceedings.mlr.press/v161/bachlechner21a.html}.

\bibitem[Buzzega et~al.(2020)Buzzega, Boschini, Porrello, Abati, and
  Calderara]{buzzega2020derpp}
Pietro Buzzega, Matteo Boschini, Angelo Porrello, Davide Abati, and Simone
  Calderara.
\newblock Dark experience for general continual learning: a strong, simple
  baseline.
\newblock In \emph{Advances in Neural Information Processing Systems
  (NeurIPS)}, 2020.

\bibitem[Caldas et~al.(2018)Caldas, Kone{\v{c}}n{\'y}, McMahan, and
  Talwalkar]{caldas2018feddropout}
Sebastian Caldas, Jakub Kone{\v{c}}n{\'y}, H.~Brendan McMahan, and Ameet
  Talwalkar.
\newblock Expanding the reach of federated learning by reducing client resource
  requirements.
\newblock \emph{arXiv preprint arXiv:1812.07210}, 2018.

\bibitem[Chaudhry et~al.(2018)Chaudhry, Dokania, Ajanthan, and
  Torr]{chaudhry2018forgetting}
Arslan Chaudhry, Puneet~K. Dokania, Thalaiyasingam Ajanthan, and Philip H.~S.
  Torr.
\newblock Riemannian walk for incremental learning: Understanding forgetting
  and intransigence.
\newblock In \emph{Proceedings of the European Conference on Computer Vision},
  pages 532--547, 2018.
\newblock URL
  \url{https://openaccess.thecvf.com/content_ECCV_2018/html/Arslan_Chaudhry__Riemannian_Walk_ECCV_2018_paper.html}.

\bibitem[Chaudhry et~al.(2019)Chaudhry, Rohrbach, Elhoseiny, Ajanthan, Dokania,
  Torr, and Ranzato]{chaudhry2019er}
Arslan Chaudhry, Marcus Rohrbach, Mohamed Elhoseiny, Thalaiyasingam Ajanthan,
  Puneet~K. Dokania, Philip H.~S. Torr, and Marc'Aurelio Ranzato.
\newblock On tiny episodic memories in continual learning.
\newblock \emph{arXiv preprint arXiv:1902.10486}, 2019.

\bibitem[De and Smith(2020)]{de2020skipinit}
Soham De and Samuel~L. Smith.
\newblock Batch normalization biases residual blocks towards the identity
  function in deep networks.
\newblock In \emph{Advances in Neural Information Processing Systems},
  volume~33, 2020.
\newblock URL
  \url{https://proceedings.neurips.cc/paper/2020/hash/e6b738eca0e6792ba8a9cbcba6c1881d-Abstract.html}.

\bibitem[Devvrit et~al.(2024)Devvrit, Kudugunta, Kusupati, Dettmers, Chen,
  Dhillon, Tsvetkov, Hajishirzi, Kakade, Farhadi, and
  Jain]{devvrit2024matformer}
Devvrit, Sneha Kudugunta, Aditya Kusupati, Tim Dettmers, Kaifeng Chen, Inderjit
  Dhillon, Yulia Tsvetkov, Hannaneh Hajishirzi, Sham Kakade, Ali Farhadi, and
  Prateek Jain.
\newblock {MatFormer}: Nested transformer for elastic inference.
\newblock In \emph{Advances in Neural Information Processing Systems},
  volume~37, 2024.
\newblock \doi{10.52202/079017-4461}.
\newblock URL
  \url{https://proceedings.neurips.cc/paper_files/paper/2024/hash/fe066022bab2a6c6a3c57032a1623c70-Abstract-Conference.html}.

\bibitem[Diao et~al.(2021)Diao, Ding, and Tarokh]{diao2021heterofl}
Enmao Diao, Jie Ding, and Vahid Tarokh.
\newblock {H}etero{FL}: Computation and communication efficient federated
  learning for heterogeneous clients.
\newblock In \emph{International Conference on Learning Representations
  (ICLR)}, 2021.

\bibitem[Dong et~al.(2022)Dong, Wang, Fang, Sun, Xu, Wang, and
  Zhu]{dong2022glfc}
Jiahua Dong, Lixu Wang, Zhen Fang, Gan Sun, Shichao Xu, Xiao Wang, and Qi~Zhu.
\newblock Federated class-incremental learning.
\newblock In \emph{Proceedings of the IEEE/CVF Conference on Computer Vision
  and Pattern Recognition (CVPR)}, 2022.

\bibitem[Fan et~al.(2020)Fan, Grave, and Joulin]{fan2020layerdrop}
Angela Fan, Edouard Grave, and Armand Joulin.
\newblock Reducing transformer depth on demand with structured dropout.
\newblock In \emph{International Conference on Learning Representations}, 2020.
\newblock URL \url{https://openreview.net/forum?id=SylO2yStDr}.

\bibitem[Hong et~al.(2022)Hong, Wang, Wang, and Zhou]{hong2022splitmix}
Junyuan Hong, Haotao Wang, Zhangyang Wang, and Jiayu Zhou.
\newblock Efficient split-mix federated learning for on-demand and in-situ
  customization.
\newblock In \emph{International Conference on Learning Representations
  (ICLR)}, 2022.

\bibitem[Horvath et~al.(2021)Horvath, Laskaridis, Almeida, Leontiadis,
  Venieris, and Lane]{horvath2021fjord}
Samuel Horvath, Stefanos Laskaridis, Mario Almeida, Ilias Leontiadis, Stylianos
  Venieris, and Nicholas Lane.
\newblock {FjORD}: Fair and accurate federated learning under heterogeneous
  targets with ordered dropout.
\newblock In \emph{Advances in Neural Information Processing Systems
  (NeurIPS)}, pages 12876--12889, 2021.

\bibitem[Hsieh et~al.(2020)Hsieh, Phanishayee, Mutlu, and
  Gibbons]{hsieh2020quagmire}
Kevin Hsieh, Amar Phanishayee, Onur Mutlu, and Phillip~B. Gibbons.
\newblock The non-{IID} data quagmire of decentralized machine learning.
\newblock In \emph{International Conference on Machine Learning (ICML)}, volume
  119 of \emph{Proceedings of Machine Learning Research}, pages 4387--4398.
  PMLR, 2020.

\bibitem[Huang et~al.(2016)Huang, Sun, Liu, Sedra, and
  Weinberger]{huang2016stochasticdepth}
Gao Huang, Yu~Sun, Zhuang Liu, Daniel Sedra, and Kilian~Q. Weinberger.
\newblock Deep networks with stochastic depth.
\newblock In \emph{European Conference on Computer Vision}, pages 646--661,
  2016.
\newblock \doi{10.1007/978-3-319-46493-0_39}.
\newblock URL
  \url{https://link.springer.com/chapter/10.1007/978-3-319-46493-0_39}.

\bibitem[Ilhan et~al.(2023)Ilhan, Su, and Liu]{ilhan2023scalefl}
Fatih Ilhan, Gong Su, and Ling Liu.
\newblock {S}cale{FL}: Resource-adaptive federated learning with heterogeneous
  clients.
\newblock In \emph{IEEE/CVF Conference on Computer Vision and Pattern
  Recognition (CVPR)}, pages 24532--24541, 2023.

\bibitem[Kim et~al.(2023)Kim, Yu, Kim, and Moon]{kim2023depthfl}
Minjae Kim, Sangyoon Yu, Suhyun Kim, and Soo-Mook Moon.
\newblock {D}epth{FL}: Depthwise federated learning for heterogeneous clients.
\newblock In \emph{International Conference on Learning Representations
  (ICLR)}, 2023.

\bibitem[Kirkpatrick et~al.(2017)Kirkpatrick, Pascanu, Rabinowitz, Veness,
  Desjardins, Rusu, Milan, Quan, Ramalho, Grabska-Barwinska, Hassabis, Clopath,
  Kumaran, and Hadsell]{kirkpatrick2017ewc}
James Kirkpatrick, Razvan Pascanu, Neil Rabinowitz, Joel Veness, Guillaume
  Desjardins, Andrei~A. Rusu, Kieran Milan, John Quan, Tiago Ramalho, Agnieszka
  Grabska-Barwinska, Demis Hassabis, Claudia Clopath, Dharshan Kumaran, and
  Raia Hadsell.
\newblock Overcoming catastrophic forgetting in neural networks.
\newblock \emph{Proceedings of the National Academy of Sciences}, 114\penalty0
  (13):\penalty0 3521--3526, 2017.
\newblock \doi{10.1073/pnas.1611835114}.

\bibitem[Kusupati et~al.(2022)Kusupati, Bhatt, Rege, Wallingford, Sinha,
  Ramanujan, Howard-Snyder, Chen, Kakade, Jain, and
  Farhadi]{kusupati2022matryoshka}
Aditya Kusupati, Gantavya Bhatt, Aniket Rege, Matthew Wallingford, Aditya
  Sinha, Vivek Ramanujan, William Howard-Snyder, Kaifeng Chen, Sham Kakade,
  Prateek Jain, and Ali Farhadi.
\newblock Matryoshka representation learning.
\newblock In \emph{Advances in Neural Information Processing Systems
  (NeurIPS)}, 2022.

\bibitem[Li et~al.(2024)Li, Li, Wang, Li, Zhong, and Zhang]{li2024refed}
Yichen Li, Qunwei Li, Haozhao Wang, Ruixuan Li, Wenliang Zhong, and Guannan
  Zhang.
\newblock Towards efficient replay in federated incremental learning.
\newblock In \emph{Proceedings of the IEEE/CVF Conference on Computer Vision
  and Pattern Recognition (CVPR)}, 2024.

\bibitem[Li and Hoiem(2016)]{li2016lwf}
Zhizhong Li and Derek Hoiem.
\newblock Learning without forgetting.
\newblock In \emph{European Conference on Computer Vision (ECCV)}, pages
  614--629, 2016.

\bibitem[Liu et~al.(2022)Liu, Wu, Wu, Wang, Lyu, Chen, and
  Xie]{liu2022inclusivefl}
Ruixuan Liu, Fangzhao Wu, Chuhan Wu, Yanlin Wang, Lingjuan Lyu, Hong Chen, and
  Xing Xie.
\newblock No one left behind: Inclusive federated learning over heterogeneous
  devices.
\newblock In \emph{Proceedings of the 28th ACM SIGKDD Conference on Knowledge
  Discovery and Data Mining}, pages 3398--3406, 2022.
\newblock \doi{10.1145/3534678.3539086}.
\newblock URL \url{https://doi.org/10.1145/3534678.3539086}.

\bibitem[Liu and Huang(2025)]{liu2025spapficl}
Youchao Liu and Dingjiang Huang.
\newblock Sparse personalized federated class-incremental learning.
\newblock \emph{Information Sciences}, 706:\penalty0 121992, 2025.
\newblock \doi{10.1016/j.ins.2025.121992}.
\newblock URL \url{https://doi.org/10.1016/j.ins.2025.121992}.

\bibitem[Moussadek et~al.(2025)Moussadek, Salami, and
  Calderara]{moussadek2025dolfin}
Omayma Moussadek, Riccardo Salami, and Simone Calderara.
\newblock {DOLFIN}: Balancing stability and plasticity in federated continual
  learning.
\newblock \emph{arXiv preprint arXiv:2510.13567}, 2025.

\bibitem[Nguyen et~al.(2025)Nguyen, Nguyen, and Pham]{stamp2025}
Minh-Duong Nguyen, Le-Tuan Nguyen, and Quoc-Viet Pham.
\newblock Improving generalization in heterogeneous federated continual
  learning via spatio-temporal gradient matching with prototypical coreset.
\newblock \emph{arXiv preprint arXiv:2506.12031}, 2025.
\newblock \doi{10.48550/arXiv.2506.12031}.
\newblock URL \url{https://arxiv.org/abs/2506.12031}.

\bibitem[Prabhu et~al.(2020)Prabhu, Torr, and Dokania]{prabhu2020gdumb}
Ameya Prabhu, Philip H.~S. Torr, and Puneet~K. Dokania.
\newblock {GDumb}: A simple approach that questions our progress in continual
  learning.
\newblock In \emph{European Conference on Computer Vision (ECCV)}, 2020.

\bibitem[Rebuffi et~al.(2017)Rebuffi, Kolesnikov, Sperl, and
  Lampert]{rebuffi2017icarl}
Sylvestre-Alvise Rebuffi, Alexander Kolesnikov, Georg Sperl, and Christoph~H.
  Lampert.
\newblock {iCaRL}: Incremental classifier and representation learning.
\newblock In \emph{IEEE Conference on Computer Vision and Pattern Recognition
  (CVPR)}, pages 5533--5542, 2017.

\bibitem[Rippel et~al.(2014)Rippel, Gelbart, and
  Adams]{rippel2014nesteddropout}
Oren Rippel, Michael Gelbart, and Ryan Adams.
\newblock Learning ordered representations with nested dropout.
\newblock In \emph{Proceedings of the 31st International Conference on Machine
  Learning}, volume~32 of \emph{Proceedings of Machine Learning Research},
  pages 1746--1754, 2014.
\newblock URL \url{https://proceedings.mlr.press/v32/rippel14.html}.

\bibitem[Touvron et~al.(2021)Touvron, Cord, Sablayrolles, Synnaeve, and
  J{\'e}gou]{touvron2021layerscale}
Hugo Touvron, Matthieu Cord, Alexandre Sablayrolles, Gabriel Synnaeve, and
  Herv{\'e} J{\'e}gou.
\newblock Going deeper with image transformers.
\newblock In \emph{Proceedings of the IEEE/CVF International Conference on
  Computer Vision}, pages 32--42, 2021.
\newblock URL
  \url{https://openaccess.thecvf.com/content/ICCV2021/html/Touvron_Going_Deeper_With_Image_Transformers_ICCV_2021_paper.html}.

\bibitem[van~de Ven and Tolias(2019)]{vandeven2019three}
Gido~M. van~de Ven and Andreas~S. Tolias.
\newblock Three scenarios for continual learning.
\newblock \emph{arXiv preprint arXiv:1904.07734}, 2019.

\bibitem[Veit et~al.(2016)Veit, Wilber, and Belongie]{veit2016resnetensembles}
Andreas Veit, Michael~J. Wilber, and Serge Belongie.
\newblock Residual networks behave like ensembles of relatively shallow
  networks.
\newblock In \emph{Advances in Neural Information Processing Systems
  (NeurIPS)}, 2016.

\bibitem[Wu et~al.(2024)Wu, Wang, Wang, Liu, Su, and Gao]{wu2024fiarse}
Feijie Wu, Xingchen Wang, Yaqing Wang, Tianci Liu, Lu~Su, and Jing Gao.
\newblock {FIARSE}: Model-heterogeneous federated learning via importance-aware
  submodel extraction.
\newblock In \emph{Advances in Neural Information Processing Systems},
  volume~37, 2024.
\newblock \doi{10.52202/079017-3672}.
\newblock URL
  \url{https://proceedings.neurips.cc/paper_files/paper/2024/hash/d17477a5ff8b7ddb49e53f9c04305da5-Abstract-Conference.html}.

\bibitem[Wu et~al.(2019)Wu, Chen, Wang, Ye, Liu, Guo, and Fu]{wu2019bic}
Yue Wu, Yinpeng Chen, Lijuan Wang, Yuancheng Ye, Zicheng Liu, Yandong Guo, and
  Yun Fu.
\newblock Large scale incremental learning.
\newblock In \emph{IEEE Conference on Computer Vision and Pattern Recognition
  (CVPR)}, pages 374--382, 2019.
\newblock BiC.

\bibitem[Wu and He(2018)]{wu2018groupnorm}
Yuxin Wu and Kaiming He.
\newblock Group normalization.
\newblock In \emph{European Conference on Computer Vision (ECCV)}, pages 3--19,
  2018.
\newblock \doi{10.1007/978-3-030-01261-8_1}.
\newblock URL
  \url{https://link.springer.com/chapter/10.1007/978-3-030-01261-8_1}.

\bibitem[Yoon et~al.(2021)Yoon, Jeong, Lee, Yang, and Hwang]{yoon2021fedweit}
Jaehong Yoon, Wonyong Jeong, Giwoong Lee, Eunho Yang, and Sung~Ju Hwang.
\newblock Federated continual learning with weighted inter-client transfer.
\newblock In \emph{International Conference on Machine Learning (ICML)}, volume
  139 of \emph{Proceedings of Machine Learning Research}, pages 12073--12086.
  PMLR, 2021.

\bibitem[Yu et~al.(2025{\natexlab{a}})Yu, Yang, Fan, Cao, Gu, Fan, and
  Yang]{fedlscl2025}
Hao Yu, Xin Yang, Boyang Fan, Xuemei Cao, Hanlin Gu, Lixin Fan, and Qiang Yang.
\newblock Large-small model collaborative framework for federated continual
  learning.
\newblock \emph{arXiv preprint arXiv:2508.09489}, 2025{\natexlab{a}}.
\newblock \doi{10.48550/arXiv.2508.09489}.
\newblock URL \url{https://arxiv.org/abs/2508.09489}.

\bibitem[Yu et~al.(2025{\natexlab{b}})Yu, Yang, Zhang, Gu, Li, Fan, and
  Yang]{fedta2025}
Hao Yu, Xin Yang, Le~Zhang, Hanlin Gu, Tianrui Li, Lixin Fan, and Qiang Yang.
\newblock Handling spatial-temporal data heterogeneity for federated continual
  learning via tail anchor.
\newblock In \emph{Proceedings of the IEEE/CVF Conference on Computer Vision
  and Pattern Recognition}, pages 4874--4883, 2025{\natexlab{b}}.
\newblock URL
  \url{https://openaccess.thecvf.com/content/CVPR2025/html/Yu_Handling_Spatial-Temporal_Data_Heterogeneity_for_Federated_Continual_Learning_via_Tail_CVPR_2025_paper.html}.

\bibitem[Yu and Huang(2019)]{yu2019usnets}
Jiahui Yu and Thomas Huang.
\newblock Universally slimmable networks and improved training techniques.
\newblock In \emph{IEEE/CVF International Conference on Computer Vision
  (ICCV)}, 2019.

\bibitem[Yu et~al.(2019)Yu, Yang, Xu, Yang, and Huang]{yu2019slimmable}
Jiahui Yu, Linjie Yang, Ning Xu, Jianchao Yang, and Thomas Huang.
\newblock Slimmable neural networks.
\newblock In \emph{International Conference on Learning Representations
  (ICLR)}, 2019.

\bibitem[Zhang et~al.(2019)Zhang, Dauphin, and Ma]{zhang2019fixup}
Hongyi Zhang, Yann~N. Dauphin, and Tengyu Ma.
\newblock Fixup initialization: Residual learning without normalization.
\newblock In \emph{International Conference on Learning Representations}, 2019.
\newblock URL \url{https://openreview.net/forum?id=H1gsz30cKX}.

\bibitem[Zhang et~al.(2023)Zhang, Dai, Wang, Xing, Chen, and Sun]{fedepth2023}
Kai Zhang, Yutong Dai, Hongyi Wang, Eric~P. Xing, Xun Chen, and Lichao Sun.
\newblock Memory-adaptive depth-wise heterogeneous federated learning.
\newblock \emph{arXiv preprint arXiv:2303.04887}, 2023.
\newblock \doi{10.48550/arXiv.2303.04887}.
\newblock URL \url{https://arxiv.org/abs/2303.04887}.

\bibitem[Zhang et~al.(2025)Zhang, Chen, Yuan, Zou, Zhuang, Jiao, Wang, and
  Yu]{feddcl2025}
Xiao Zhang, Zengzhe Chen, Yuan Yuan, Yifei Zou, Fuzhen Zhuang, Wenyu Jiao, Yuke
  Wang, and Dongxiao Yu.
\newblock Data-free continual learning of server models in model-heterogeneous
  cloud-device collaboration.
\newblock \emph{arXiv preprint arXiv:2509.25977}, 2025.
\newblock \doi{10.48550/arXiv.2509.25977}.
\newblock URL \url{https://arxiv.org/abs/2509.25977}.

\end{thebibliography}
